# The Twin-System Approach as One Generic Solution for XAI: An Overview of ANN-CBR Twins for Explaining Deep Learning

**Mark T. Keane**[1,2,3], **Eoin M. Kenny**[1,3]

[1]School of Computer Science, University College Dublin, Dublin, Ireland
[2]Insight Centre for Data Analytics, Dublin, Ireland
[3]VistaMilk SFI Research Centre, Ireland
mark.keane@ucd.ie, eoin.kenny@insight-centre.org

## Abstract

The notion of *twin-systems* is proposed to address the eXplainable AI (XAI) problem, where an uninterpretable black-box system is mapped to a white-box "twin" that is more interpretable. In this short paper, we overview very recent work that advances a generic solution to the XAI problem, the so-called *twin-system approach*. The most popular twinning in the literature is that between an Artificial Neural Networks (ANN[1]) as a black box and Case Based Reasoning (CBR) system as a white-box, where the latter acts as an interpretable proxy for the former. We outline how recent work reviving this idea has applied it to deep learning methods. Furthermore, we detail the many fruitful directions in which this work may be taken; such as, determining the most (i) accurate feature-weighting methods to be used, (ii) appropriate deployments for explanatory cases, (iii) useful cases of explanatory value to users.

## 1 Introduction

The problem of eXplainable AI (XAI) has unleashed a fairly unprecedented tsunami of research in the last few years, as almost every major AI/ML conference and workshop has targeted the problem, thematically (e.g., NIPS-16, IJCAI-17, IJCAI/ECAI-18, IJCNN-17, ICCBR-18, ICCBR-19, IJCAI-19 and the present workshop), along with the emergence of conferences dedicated solely to it (FAT-ML, FAT*19; see [Adadi and Berrada, 2018]). Hence, a focus on the problem is no longer the primary issue of concern, it is perhaps the question of whether we are making progress on the XAI problem and fashioning adequate solutions to it. In this light, the present paper overviews a very recent literature that proposes a generic solution to one aspect of the XAI problem -- namely, that of *post-hoc explanation by example* – called the *twin system approach* (see Keane & Kenny, [2019]; Kenny and Keane, [2019]; Kenny et al., [2019]).

Some recent reviews of the literature have begun to comment on the state of current XAI solutions [Guidotti *et al.*, 2018; Pedreschi *et al.*, 2019]. For instance, Pedreschi *et al.* [2019] have argued that current research is throwing up fragmentary solutions that appear to lack generality; to quote them, they say "the state of the art to date still exhibits *ad-hoc*, scattered results, mostly hardwired to specific models…[and]… a widely applicable, systematic approach has not emerged yet". The issue for XAI now appears to be the formulation of a systematic, general framework to bring the literature together and focus future efforts. The *twin-system approach* is proposed as one such solution, where an uninterpretable black-box system (typically, an ANN) is mapped to a white-box "twin" that is more interpretable (typically, a CBR system; see [Aamodt and Plaza, 1994; Mantaras *et al.*, 2006]).

Simultaneously, many reviews have tried to provide taxonomies to cut up and organize the XAI problem. Over a decade ago, reflecting ideas from Philosophy, Sørmo *et al.* [2005] reported the distinction between explaining how a system might reach some answer (what they call *transparency*) and explaining why the answer is good (*justification*; see also Tintarev and Masthoff [2007], Nunes and Jannach [2017]). Recently, this distinction has been echoed by dividing *interpretability* into (i) *transparency* (or *simulatability*) which tries to reflect *how* the AI system produced its outputs, and (ii) *post-hoc interpretability* which is more about *why* the AI reached its outputs, providing some after-the-fact rationale/evidence for system outputs [Lipton, 2018; Biran and Cotton, 2017]. Indeed, *post-hoc* explanation has been further sub-divided into the use of (i) textual explanations of system outputs, (ii) visualizations of learned representations and/or models [Erhan *et al.*, 2009; Zeiler and Fergus, 2014], and (iii) explanations by example [Lipton, 2018]. With respect to the *interpretability* of Deep Neural Networks (DNNs), Giblin *et al.* [8] propose that they may be explained using a *proxy model* "which behaves similarly to the original model, but in a way that is easier to explain, or by creating a *saliency map* to highlight a small portion of the computation that is relevant" (p. 3); they identify *linear proxy*

---

[1] Note, we use the term Artificial Neural Network (ANN) as a generic umbrella term for classic Multi-layered Perceptron (MLP) architectures and newer Deep Neural Network (DNN) methods.

*models* (e.g., LIME [Ribeiro *et al*., 2016]) and *decision trees* as common options for such proxy models. As we shall see, the twin-system approach overviewed here, has elements of post-hoc explanation-by-example and proxy modelling, though it can also accommodate saliency maps (perhaps, suggesting that these taxonomic cuts are not as deep as they first appear).

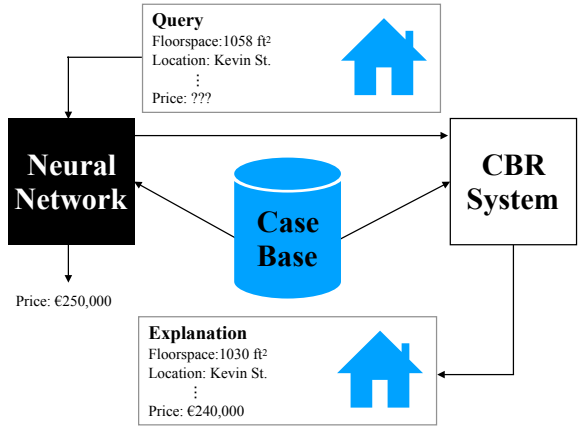

Figure 1: A simple ANN-CBR twin-system (adapted from [Kenny and Keane, 2019]); a query-case posed to an ANN gives an accurate, but unexplained, prediction for a house price. The ANN is twinned with the CBR system (both use the same dataset), and its feature-weights are analyzed and used in the CBR system to retrieve nearest-neighbor cases to explain the ANN's prediction.

The present paper brings these strands together to overview a collection of very recent work on what has been called, the *twin-system approach,* to provide a perspective not available in each of these individual papers [Keane & Kenny, 2019; Kenny and Keane, 2019; Kenny *et al.,* 2019]. This approach is proposed as a generic solution to one solution to the XAI problem; namely, the use of *post-hoc* explanation-by-example. Twin-systems are hybrid systems, with specific properties, and may involve "twinnings" between varied AI techniques, but ANN-CBR twinning has, perhaps, been the most used (so called ANN-CBR twins). For example, consider an ANN that accurately predicts house prices given different feature-descriptions of houses (e.g., *size*, *location*, *no-of-rooms*), but like all ANNs is opaque and, thus, cannot explain its predictions easily. If we twin this ANN with a CBR system, using the same dataset, extract the ANN's feature-weights and apply them in the CBR-system to retrieve neighboring cases to the query-case, then we can use the latter to explain the predictions of the former (see Fig. 1). One of the issues in this area is how best to characterize the feature weights of ANNs; that is, which feature-weighting technique has the highest fidelity to the ANN's function (which, as we shall see, leads to a consideration of other proxy methods, such as LIME).

In providing an overview of this very recent work, we attempt to show that it qualifies as a general approach to one class of solution to the XAI problem as it (i) can be applied to a diverse set of ANN architectures from traditional multi-layer perceptron networks (MLPs) to more recent DNNs, (ii) it has been applied to a wide variety of domains (from traditional tabular data, such as credit ratings, to the image-datasets used in DNNs), and (iii) it provides a general framework for twinning different techniques.

In the remainder of this paper, we first define the twin-system idea (see Section 2), before sketching some of the main findings from ANN-CBR twinnings in the past and present literature (see Section 3); finally, we consider some of the wider implications of the approach (see Section 4).

## 2 A Definition of Twin-Systems

Twin-systems are predicated on the idea that if some description/characterization/representation of an opaque black-box model (such as an ANN) can be derived and mapped to a more-interpretable white-box "twin", then the former can be explained by the latter. Though it has not always be recognized as such, one of the most researched examples of this twinning is between opaque ANNs and more transparent CBR systems (i.e., ANN-CBR twins). CBR systems are known for their intuitive reasoning method (i.e., similarity between cases) and their use of explanation by example/case/precedent [Aamodt and Plaza, 1994; Mantaras *et al*., 2006; Leake and McSherry, 2005]. The key aspect of twinning is that both systems are applied to the same dataset and the feature-weights of the ANN are used in the CBR system's retrieval step (typically using *k*-NN) to retrieve explanatory cases. Keane and Kenny [2019] have defined ANN-CBR twins to involve:

- *Two Techniques*. A hybrid system where an ANN (MLP or DNN) and a CBR technique (notably, *k*-NN) are combined to meet the system requirements of accuracy and interpretability.
- *Separate Modules.* Where these techniques are run as separate, independent modules, "side-by-side".
- *Common Dataset.* Both techniques are applied to the same dataset (i.e., twinned by this common usage).
- *Feature-Weight Mapping.* Where some description of the ANN's functionality, cast as its *feature-weights*, reflecting what the ANN has learned, is mapped to the *k*-NN retrieval step of the CBR-system.
- *Bipartite Division of Labor*. There is a bipartite division of labor between the ANN and CBR modules, where the former delivers prediction-accuracy, and the latter provides interpretability by explaining the ANN's outputs (in classification or regression), using example cases.

They applied this definition to the literature on "hybrid systems involving explanation" covering the last 25+ years and found a fragmented literature echoing the twin-system idea (in all but name). However, less than a handful of papers seriously addressed the core issue; namely, what was the best feature-weighting scheme to use to describe ANNs when they were being explained by these CBR systems.

## 3 Twin Systems: What is Past or Passing…

One of the characteristics of recent popular XAI reviews is the tendency to overlook the past literature (see e.g., Lipton, [2018], Doshi-Velez and Kim, [2017], Gilpin *et al.*, [2019]), when that literature often sheds significant light on current problems, even though techniques have advanced and changed. In response, Keane and Kenny [2019] carried out a systematic review of the literature to find papers that met the twin-systems definition. This review turned up 21 previous ANN-CBR twin-systems and some significant work on the impact of different feature-weighing schemes. Their conclusions are summarized here looking back at historical results (see Section 3.1), more recent findings (see Section 3.2) and some future directions of relevance to Deep Learning techniques (see Section 3.3; see Keane and Kenny [2019] for details).

### 3.1 Historical Findings: What is Past…

The historical work on twin systems divides into a late-1990s collection of papers by a Korean group at KAIST (e.g., [Park *et al*., 2004]) and some mid-2000s work by an Irish CBR group (e.g., [Cunningham *et al*., 2003; Nugent and Cunningham, 2005; Doyle *et al*., 2004]).

The early Korean work addressed different feature-weighting schemes that described MLPs (e.g., *sensitivity*, *activity*, *relevance* and *saliency*) to find the best scheme to apply to the case-features of the *k*-NN for case-based explanation. They tested a wide range of datasets and found that *sensitivity* and *activity* tended to do best for many datasets [Shin and Park, 1999; Shin *et al.,* 2000; Im and Park, 2007]. For the most part, this research does not involve controlled user-tests to determine the explanatory value of case-based explanations, but the authors did report that expert users were satisfied [Im and Park, 2007]. This Korean group also advanced an important distinction between *global* and *local* feature-weighting schemes, where (i) *global methods* take the input space as isotropic, deriving a single ubiquitous feature-weight vector for the entire domain, and (ii) *local methods* were a specific set of weights for each query case is found [Park *et al.*, 2004].

Independently of the Korean group, in the mid-2000s an Irish group explored another local feature-weighting method [Nugent and Cunningham, 2005]. Nugent and Cunningham [2005] built an artificial local dataset around a given query by systematically perturbing the features of it before querying labels for these cases in the MLP. They then proceeded to build a local linear model (similar to LIME) using this new local dataset; the coefficients of the linear model were then used for weighting *k*-NN searches to retrieve explanatory cases. Significantly, this group also performed a number of controlled user tests to see whether the retrieved cases had any explanatory value, finding that provision of cases improved user satisfaction of the system [Nugent *et al*., 2009].

### 3.2 More Recent Findings: Past, or Passing…

Keane and Kenny's [2019] survey found that much of this early work has either not been recgnised in recent reviews or has been sporadically cited. However, they also found some recent papers that attempt to take up this early work and extend it. For example, Biswas *et al*. [2017] revisited the Korean work to improve it for unbalanced datasets.

Furthermore, Kenny and Keane [2019] revisit the techniques tested in this early work and compare them to more recent methods (e.g., DeepLIFT [Shrikumar *et al.*, 2017] and LIME [Riberio *et al.*, 2016]) to see whether they work well in a wide variety of ANN-CBR twins.

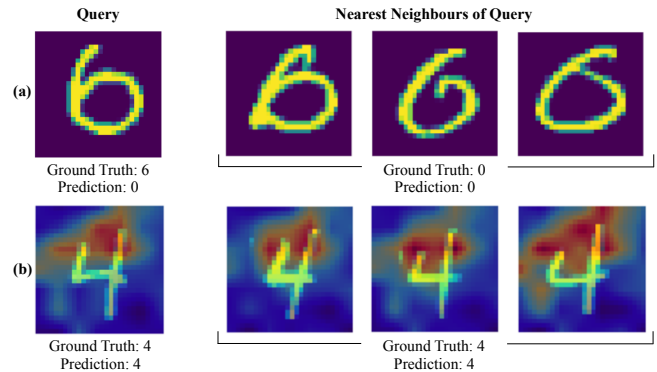

Figure 2: The CNN classifies an image of the number "6" as "0". The explanation is given as nearest neighbors from the case base. In (a) because the model was trained on data which was labelled as a "0" but looks like a "6" and, consequently, miss-classifies the query (adapted form Kenny and Keane [2019]), (b) the image of a "4" is correctly identified, and the feature-activation map (FAM) shows that a critical aspect of this prediction is the "link" across the top of the number (which, if present, might make it a "9").

### 3.3 Findings on DNN Twins: Passing, or to Come…

Finally, most previous twin-system work has tended to examine traditional neural networks (i.e., MLPs), rather than more recent DNNs (for rare potential exceptions see [Chen *et al.,* 2018] and [Li *et al.,* 2017]). However, Kenny and Keane [2019] have extended tests of feature-weighting schemes to convolutional neural networks (CNNs). Specifically, they have shown that DeepLIFT provides the best performance for twin-systems whether they be MLP-CBR or CNN-CBR twins.

For example, Fig. 2 shows a CNN example using the MNIST dataset where DeepLIFT was used to capture the feature-weightings of the CNN; these weights were then applied to find nearest neighbors in a twinned CBR system for a given query-prediction pair. On the face of it, the CNN appears to have made an egregious error, it classifies the query as a "0" when the ground-truth identifies it as a "6". However, when one examines the explanatory, nearest-neighbor cases that were used to train the CNN, we can see that "on the basis of its experience" the CNN has reasonably identified the query as a "0", because it received training examples of zeros that look very like cursive "6s". This result highlights just one way in which twin systems can find explanatory cases that can help interpret DNNs. Similarly, with the addition of a FAM (i.e., a feature activation map showing the most positively contributing feature in a CNN classification) it is possible to show the most discriminating feature in the prediction made. Similar findings occur in other image-datasets, such as, the CIFAR-10 dataset (see Fig. 3).

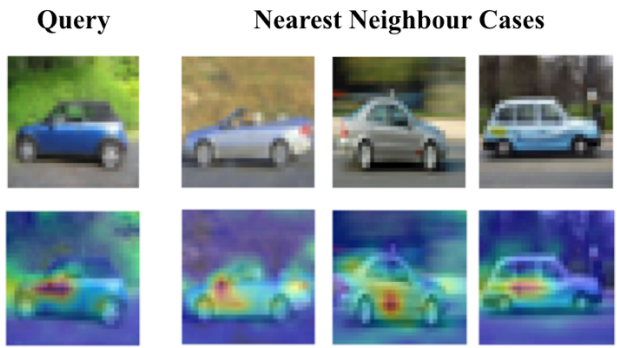

Figure 3: Adapted from Kenny and Keane [2019]: The explanation shows instances from the case base with the most similar features to the query. The most discriminating feature (highlighted in red) is shown to be consistent across the FAMs. Thus, we understand (with assurance) the primary reason for the CNN's classification.

## 4. Issues and Future Directions

Taken together, this overview of twin-systems, regarding MLPs and DNNs, shows that as an approach has a wide-ranging, and potentially generic applicability to the XAI problem. The work shows that different feature-weighting schemes (whether they be global or local) can be applied to deliver post-hoc explanations-by-example in which, in one sense, the CBR system becomes an explanatory proxy-system for the ANN. Accordingly, there are many future directions in which this work could be taken. At least five present themselves.

First, this work shows that there is much more research to be done on assessing different feature-weighting methods for use in ANN-CBR twins; more generally, it suggests the issue of between-system mapping should be established as an active and identifiable research topic of interest. One product of such a research effort would be some insight into what works best for different domains, different ANN-techniques and different use-cases.

Second, a lot more can be done on what cases are presented to explain these ANNs. We have just reported examples where supporting nearest-neighbor cases are used, but there are many other options; such as, counterfactual cases, *a fortori* cases [Nugent *et al*., 2009] or different numbers/collections of positive and counter cases (e.g., Are three nearest-neighbors better than just one?).

Third, it would be good to consider advancing the twin-system analysis to other Deep Learning techniques, beyond CNNs; for example, sequence models such as GRUs/LSTMs, and generative models such as GANs in their numerous variations. Gaining insight into how these models operate with cases is an unexamined area of research.

Fourth, the vexed question of user-evaluation remains. Keane and Kenny [2019] found that <1% of case-based explanation papers reported any type of controlled user study. Most of the deep learning literature on XAI similarly echews evaluation. So, a significant program of user-testing is required. For example, there are clearly limits to the use of cases as explanations, that can be meaningfully understood by users based on their expertise of a domain or limited-capacity processing [Sørmo *et al*., 2005]. We need to understand these limits and possible workarounds (e.g., if a case has 30+ features, a workable solution may be to use a subset of the 5-most-predictive features in explanatory cases).

Finally, this current work just looks at one twinning option – between ANNs and CBRs – to provide a solution to the XAI problem. There is a cornucopia of hybrid systems that could be explored to provide similar solutions. In short, there is a lot to keep us all very busy for the immediate future.

## Acknowledgements

This publication has emanated from research conducted with the financial support of (i) Science Foundation Ireland (SFI) to the *Insight Centre for Data Analytics* under Grant Number 12/RC/2289 and (ii) SFI and the Department of Agriculture, Food and Marine on behalf of the Government of Ireland to the *VistaMilk SFI Research Centre* under Grant Number 16/RC/3835.